# Handling Climate Change Using Counterfactuals: Using Counterfactuals in Data Augmentation to Predict Crop Growth in an Uncertain Climate Future

Mohammed Temraz[1,2], Eoin M. Kenny[2],
Elodie Ruelle[2,3], Laurence Shalloo[2,3], Barry Smyth[1], and Mark T. Keane[1,2]

[1] Insight Centre for Data Analytics, University College Dublin, Dublin, Ireland
[2] VistaMilk SFI Research Centre, University College Dublin, Dublin, Ireland
[3] Teagasc, Animal and Grassland Research, Fermoy, Ireland
{mohammed.temraz, eoin.kenny}@ucdconnect.ie
{elodie.ruelle, laurence.shalloo}@teagasc.ie
{barry.smyth, mark.keane}@ucd.ie

**Abstract.** Climate change poses a major challenge to humanity, especially in its impact on agriculture, a challenge that a responsible AI should meet. In this paper, we examine a CBR system (PBI-CBR) designed to aid sustainable dairy farming by supporting grassland management, through accurate crop growth prediction. As climate changes, PBI-CBR's historical cases become less useful in predicting future grass growth. Hence, we extend PBI-CBR using data augmentation, to specifically handle disruptive climate events, using a counterfactual method (from XAI). Study 1 shows that historical, extreme climate-events (climate outlier cases) tend to be used by PBI-CBR to predict grass growth during climate disrupted periods. Study 2 shows that synthetic outliers, generated as counterfactuals on a outlier-boundary, improve the predictive accuracy of PBI-CBR, during the drought of 2018. This study also shows that an instance-based counterfactual method does better than a benchmark, constraint-guided method.

## 1 Introduction

Climate change is arguably the single, biggest *challenge* facing the world today. The United Nations "AI for Good" platform promotes AI technologies to meet this challenge and the UN's Sustainability Goals [1]. But, how can we predict an uncertain future using historical data that may no longer apply; how can we build predictive systems that can handle the "concept drift" created by climate change, a drift that may make past training-data irrelevant. In this paper, we explore one attempt to meet such challenges in supporting a sustainable smart agriculture. We show how an AI system, PBI-CBR [2, 3], that aids dairy farmers in sustainable grass management, can better handle crop growth prediction in the face of disruptive climate events. Specifically, we explore the novel use of counterfactual techniques to augment training data, to improve future predictions during disruptive climate-events. The intuition is that a case-based counterfactual technique [4] can generate new cases by adapting historical cases to better handle climate change; these methods "re-combine" historical cases to produce new synthetic cases that are "offset" from past cases to better predict the future. This counterfactual technique is tested against actual grass-growth data for 2018, in Ireland, a year of disrupted weather, causing a forage crisis across the dairy sector in Europe [5].

In the next section, we detail the grass-growth prediction problem and the case-based reasoning (CBR) system developed to handle it for farmers (PBI-CBR [2, 3] see section 1.1). Then, we consider the relevant literature from CBR, counterfactual techniques in XAI and the novel use of counterfactuals in data augmentation (see section 1.2). Finally, we close this introduction by considering the research questions addressed and the novelties that arise (see section 1.3). We then report two major studies that: (a) determine how PBI-CBR currently handles the prediction of climate disruptive events (such as those in 2018; see section 2), (b) comparatively test the PBI-CBR system using two different counterfactual methods -- an instance-guided and constraint-guided one -- that generate synthetic cases differently for this prediction problem (see section 3).

### 1.1 The Problem: Grass Growth Prediction for Sustainable Dairy Farming:

While some climate activists have argued that many sectors of agriculture should simply be abandoned -- as humanity moves from a meat-based diet to a vegetarian (or indeed vegan) one -- the short-term feasibility of such radical changes is questionable. *Agroecology* may be more feasible, where farming systems are changed to embrace more sustainable practices [6]. In the dairy sector, such a move could be achieved by adopting *pasture-based dairy systems* where animals are predominantly fed on grass outdoors (i.e., on pastures) rather than on meal/supplements indoors [6]; where possible, such pasture-based systems have lower carbon costs (e.g., feed is not transported over long distances), and grassland can also be used as a carbon sink. However, such agricultural practices hinge on the development of a precision agriculture to support sustainability; in the dairy sector, one initiative relies on the accurate prediction of grass growth to help farmers estimate feed budgets for dairy herds [6-10].

**Grass Budgeting & Sustainability.** Accurate grass budgeting sits at the heart of this sustainable, dairy alternative which, in turn, hinges on farmers accurately predicting the grass growth on their farm in coming weeks [2, 8, 9]. When grass growth is predicted accurately the dairy farmer can (i) improve grass utilization, thus reducing reliance on meal/supplements (reducing the carbon costs), (ii) reduce fertilizer use (and potential nitrate pollution), and (iii) extend of the grazing season (reducing greenhouse gases, see [7]). The Irish dairy sector mainly operates a pasture-based system with well-defined sustainability goals [7]. To support these efforts, an online grassland management system supports farmers in this task, the PastureBase Ireland system [9].

**PastureBase Ireland.** Since 2013, Ireland's national agricultural research organization, *Teagasc*, have provided PastureBase Ireland (PBI, https://pasturebase.teagasc.ie) as a grassland management system for Irish dairy farmers [9]. PBI has 6,000+ users of the ~18,000 dairy farmers in Ireland. The PBI database has weekly records of *grass covers* for individual farms from 2013 to the present (here, a *cover* refers to the amount of grass available from each paddock/field of a farm on a given day). The current PBI system provides a farmer with a model of their farm (i.e., the paddock sizes) and the current herd-size, to help them estimate feed budgets for a week ahead. At present in PBI, grass growth is calculated by comparing the grass cover of the current week with the previous week's cover. In the future, PBI will provide predictions of grass growth; traditionally, using *mechanistic models* such as the *Moorepark St. Gilles Grass Growth* model (MoSt [8]). Currently, these models make region-level predictions based on weather and farm variables and farm-level predictions for selected farms. The present paper is a collaboration exploring AI techniques, PBI-CBR, in this problem domain [2].

**Predicting Grass Growth Using PBI-CBR**. PBI-CBR [2, 3] applies CBR to grass-growth prediction using historical data from the PBI system, that has been entered by farmers about their own farms; this data has cases recording the time-of-year, farm-id, current-grass-cover (i.e., dried grass biomass above 4cm grass height) and 3 weather parameters (i.e., rainfall, temperature, and solar-radiation; see Figure 1). PBI-CBR uses a *k*-NN to predict grass-growth-rates in a following week using its cases. However, the historical data is very noisy; some cases have missing data, different farms report different numbers of cases, and some are manifestly incorrect (e.g., impossible growth rates from data-entry errors). PBI-CBR cleans this data using a novel method – called *Bayesian Case-Exclusion* – where cases that are predictive-outliers were excluded using a separate gold-standard distribution of grass growth [8]. This cleaned case-base spanning several years (2013-2016) makes accurate predictions for grass growth in future years (optimally, for *k*=25-40) as well providing *post-hoc* explanatory cases from the same/similar farm. In this paper, we examine PBI-CBR's grass-growth predictions for atypical, disruptive climate events. For instance, the summer of 2018 was unusually hot with low rainfall across Europe. Grass tends to grow faster as temperatures rise (up to 25°C), but the absence of soil-water can interrupt growth and lead to burnt plants (at >30°C). So, in the Irish summer of 2018, when grass-growth rates typically are at their highest, growth fell back to near zero causing a feed crisis in the dairy sector.

```
----------------------------------------
              Normal Case
----------------------------------------
Farm ID: 1
Cover CoverID: 53210
Cover Date: 2017-08-03
GrowthRateFarm: 73.0
Week: 31
Month: 8
----------------------------------------
Parameter    |  Actual value | The mean
----------------------------------------
Rain               1.8           2.2
Temperature        18.8          18.9
Solar Radiation    1483.0        1417.7
----------------------------------------
```

```
----------------------------------------
              Extreme Case
----------------------------------------
Farm ID: 21
Cover CoverID: 4957
Cover Date: 2014-08-02
GrowthRateFarm: 9.0
Week: 31
Month: 8
----------------------------------------
Parameter    |  Actual value | The mean
----------------------------------------
Rain               14.1          2.2
Temperature        13.9          18.9
Solar Radiation    417.0         1417.7
----------------------------------------
```

**Fig. 1.** Examples of PBI-CBR's cases from two different farms for week-31 in 2017 and 2014; a "normal" case where weather values are close the mean and an "extreme" outlier case where rain is very high and solar radiation low relative to the mean (i.e., a climate-disrupted event)

## 1.2 Related Work: Counterfactuals from XAI to Data Augmentation

The main focus of the paper is the use of counterfactual methods for data augmentation as a solution to improving grass growth prediction in the context of climate change. However, to date, counterfactual methods are mainly used in explainable AI (XAI) rather than in data augmentation (for reviews see [11, 12]). In example-based post-hoc explanation strategies, counterfactual explanations have become very popular and are argued to be superior to factual explanations [13]. Imagine you have applied to an automated AI system for a loan and are refused, a *counterfactual explanation* might say

"if you requested a loan that was 10% lower, you would have got the loan". In the last two years, counterfactual methods have received huge attention in XAI. We review this XAI work and the few papers that consider using counterfactuals for data augmentation.

**Counterfactual Generation in XAI**. In CBR, counterfactuals have been traditionally cast as Nearest Unlike Neighbours (NUNs; 14-16); namely, the closest neighbouring case to a test case, just over a decision boundary in the dataset. Keane and Smyth [4] re-christened NUNs as *native counterfactuals*, to distinguish them from the *synthetic counterfactuals* generated by current XAI counterfactual methods. Wachter et al.'s [17] seminal paper cast synthetic counterfactual generation as a constraint optimization problem using gradient descent over a space of blindly-perturbed datapoints; using a loss function to find the "best counterfactual", balancing the proximity of the counterfactual case to the test case against its closeness to the decision boundary. So, this method aims to generate the "closest possible world" to the test case, in which the counterfactual case is minimally different and sparse (i.e., there few feature differences between test and counterfactual). In the XAI literature, this method has been extensively used and extended with additional constraints (diversity, causality, feature-importance) and other generative methods (such as, using genetic algorithms, GANs, VARs; see [12] for a full review). Mothilal et al.'s [18] *Diverse Counterfactual Explanations* (DiCE) extends this method to include *diversity* constraints; so that for given $p$, the set of counterfactuals produced minimizes the distance within the set, while maximizing the range of features changed across the set. DiCE generates counterfactuals that are valid, diverse, and sparse. Interestingly, [18] also proposed the notion of "substitutability" as an evaluation method for counterfactual XAI; namely, that if a set of generated set of counterfactuals were good, one could substitute them for the original dataset in prediction. In the present tests, we use DiCE as it has become a *defacto* benchmark for tests of counterfactual generation (e.g., see [19]).

However, Keane and Smyth [4] proposed a very different *case-based counterfactual method* that exploits known counterfactual relationships in the dataset. Their *instance-guided method* finds the test case's nearest-neighbour that takes part in a so-called *explanation case (xc)*. An explanation case is a pair of mutually-counterfactual cases which differ by at most 2 features. The test case and the counterfactual case from this nearest *xc* are used to generate a new "good" counterfactual for the test case, by combining the test-case features with the (at most) 2 difference-features from the *xc's* counterfactual. In the loan scenario, imagine historical cases that form a native counterfactual about customer-A who was *refused* a *\$5k* loan (a female accountant earning \$50k a year, who is *2.5 years* in her current job) and customer-B who was *granted* a *\$4k* loan (a female accountant on \$50k a year, who is *3 years* in her current job). Assume customer-C (a female accountant earning \$50K and 2.5 years in her job) has also been *refused* a \$5k loan. In this scenario, the native counterfactual suggests generated counterfactual explaining "if you wait 6-months and re-apply for a lower loan (of say \$4k), you will be *granted* the loan". So, customer-C's nearest neighbour in the dataset is customer-A (who has the same refusal outcome), but because there is a close counterfactual case, customer-B (with a different outcome), a counterfactual scenario for customer-C can be generated, using the difference-features found (time-in-job, loan-requested). In Study 2 reported here, we this method is used in data augmentation tests and compared to DiCE (see section 3.1 for a full algorithmic description).

**Counterfactual in Data Augmentation**. Beyond XAI, our hypothesis is that counterfactual methods could also play a role in data augmentation, that generated synthetic counterfactual cases could improve the predictive accuracy of a model. Though there are now 100s of papers on counterfactuals in XAI, only a handful of papers consider their use in data augmentation [22-25]. Recall, that Mothilal et al. [18] proposed *substitutability* as a way to evaluate counterfactual XAI methods; namely, that a good set of generated counterfactuals should be able to substitute for the original dataset. Hasan [22] explicitly tested this idea, using DiCE, to determine if an augmented dataset using DiCE's counterfactuals could act as a proxy dataset; however, the improvements found were minimal. [23] consider the problem of *dataset shift*, where there is a divergence between the context in which a model was trained and tested; they use the notion of "counterfactual risk" to diagnose this problem using causal models. However, this work does not use the XAI counterfactual methods that have been extensively tested; hence, this work's status, reproducibility and/or generality is unclear. So, in the current tests, we use two proven counterfactual methods from the XAI-literature (i.e., [18], [4]).

### 1.3 Research Questions & Novelties

In this paper, we test whether the counterfactual methods, developed in XAI, can be applied to the challenging concept-drift problems associated with climate-change; specifically, in the context of grass growth prediction for dairy farmers (using the PBI-CBR model). So, we determine whether generated synthetic, counterfactual cases can be used in improve prediction during periods of climate disruption (focusing on 2018). However, before we can consider whether counterfactual data augmentation might work, there are several prior steps that need to be considered. First, we need to understand how the PBI-CBR model currently handles grass-growth prediction when it encounters climate-disrupted events (as test cases); a reasonable hypothesis might be that it uses historical-cases capturing past climate-disruptive events. However, this begs the non-trivial question of how one might define "past climate-disruptive events". Hence, we perform two major studies, one that aims to understand how PBI-CBR actually predicts grass growth for climate-disruptive events (Study 1; section 2) and one that comparatively tests whether counterfactual data augmentation methods can improve PBI-CBR's performance on such climate-disruptive events (Study 2; section 3). So, these studies aim to answer 3 research questions:

*RQ1*: How does PBI-CBR currently handle grass growth prediction involving climate-disruptive events (see section 2)?
*RQ2*: Can PBI-CBR's prediction of climate-disruptive events be improved by counterfactual data-augmentation methods (section 3)?
*RQ3*: And, if prediction is improved by counterfactuals, which counterfactual methods work best (section 3)?

As we shall see, several significant novelties arise from the answers to these research questions (see section 4). In the following sections, we describe the studies carried out.

## 2 Study 1: Predicting Climate Disruption with PBI-CBR

In this first study, to answer RQ1, we analyze PBI-CBR's grass-growth predictions when it encounters climate disruptive events. However, before we can assess its performance, we need to define cases that potentially reflect climate-disruptive events (see section 2.1). Then, armed with this definition, we perform two experiments. The first experiment determines whether historical extreme-climate cases in the PBI-CBR case-base tend to be used to predict growth rates when extreme-climate test-cases are encountered (using 2018 as a test year; see Expt. 1a in section 2.2). This may seem like an obvious test but it is not. The grass-growth dataset is very noisy, as the cases come from end-user data-entry on the PBI website; while PBI-CBR automatically cleans the original dataset, it is still not clear whether "outlier" cases are "true outliers" representing actual extreme-weather events on a farm or just invalid data-points created by data-entry errors (e.g., in temperature or growth data). The second experiment considers the effects of varying *k* in the model on these results; see Expt. 1b in section 2.3).

Both experiments used PBI dataset drawn from 6,000+ farms over 6 years 2013-2018 (N=70,091) [1]; divided as follows 2013 (N=5,205), 2014 (N=6,852), 2015 (N=9,695), 2016 (N=14,777), 2017 (N=18,611), 2018 (N=14,951). In general, the number of cases increases each year as more farmers adopted the PBI system. So, in the both experiments, the years 2013-2016 (N=36,529) were used as training data and 2018 (N=14,951) was used as the test data (2017 is run too but not reported); 2018 had many extreme climate events with high-temperatures, high solar-radiation and low summer rainfall that caused a feed-crisis for the sector (as grass growth was inhibited by low soil moisture and solar radiation damage). As such, it is real-life, test-case of the climate challenges now facing agriculture. However, we first need to define which cases are likely to be ones reflecting extreme-climate events.

### 2.1 Defining a Class Boundary for Climate Outlier Cases

To run our tests on PBI-CBR we need some definition of which cases might reflect extreme-weather events (n.b., extreme values could just be data-entry errors). Here, we used a statistical approach to define, what we call, *climate outlier cases;* that is, cases that appear to capture extreme weather events by virtue of having high/low extreme values for either temperature, rainfall, or solar radiation. As weather data follows a normal distribution $[X \sim (\mu, \sigma^2)]$ for each week, we defined *climate outliers* as cases with values that are >2 standard deviations above/below the mean for a given week. So, this filter was applied to all the cases for a given week (e.g., week 12) aggregated over all the years (2013-18) in the dataset (an in-year weekly-average produces broadly similar results). More formally, for weather parameters high/low outliers are defined as:

$$\textit{High Outliers} = \; X_i > \mu + 2\sigma$$
$$\textit{Low Outliers} = \; X_i < \mu - 2\sigma$$

where $X_i$ is an observation, $\mu$ and $\sigma$ are the mean and standard deviation for a given week. Figure 1 shows two sample cases, a "normal" farm case with typical weather features for the week-31 of 2017 and a "extreme" climate-outlier case for the same week of 2014. Figure 2a shows the distribution of *temperature* values for each week across 2013-2018 (with box plots) and Figure 2b shows the high and low outliers found for each week in this combined dataset. Note, how there are many high-temperature outliers in summer weeks and many low-temperature ones in winter weeks.

[1] Note, this is after pre-processing to remove noisy cases (originally, N=138,970).

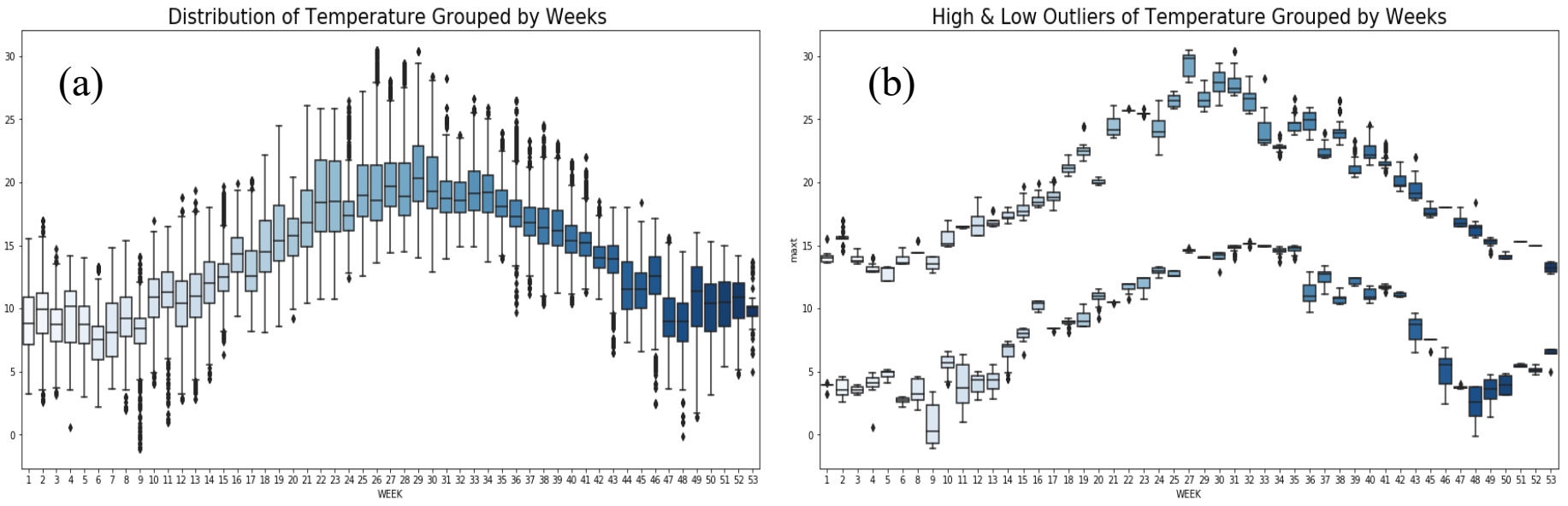

**Fig. 2.** The distribution of temperature values (with box plots) (a) for all cases by weeks of the year (from 2013-2018) with (b) high and low outliers separated out.

When we remove these climate outliers from the overall PBI-CBR dataset (N=70,091) for 2013-2018, we find 7,324 *unique outliers*[2]. Most climate outliers reflect rainfall extremes (44%, N=3,500), with others reflecting extremes of temperature (38%, N=2,997) and solar radiation (18%, N=1,414). The percentage of outlier cases in each year is fairly constant, though frequencies increase across years (in-year %'s shown): 2013 (16%, N=836), 2014 (10%, N=707), 2015 (10%, N=1,008), 2016 (9%, N=1,259), 2017 (10%, N=1,778), 2018 (12%, N=1,736). In these experiments, we used 2013-2016 as the training set, testing it mainly against 2018 (we found equivalent results for 2017, though as it was more "normal", the effects were less pronounced). So, in these experiments the 2013-16 PBI-CBR case-base had N=36,529 cases, when all cases are included, and N=32,719 cases when the climate outliers were excluded.

**Table 1.** Frequencies of training outliers used to predict test outliers in 2018 (Expt.1a)

| | Training (*Outliers*) | Training (*Non-Outliers*) |
|---|---|---|
| *Test-Outliers* (N=1,736) | 1,534 *(88.4%)* | 202 *(11.6%)* |
| *Test-Non-Outliers* (N=1,248) | 144 (*11.5%)* | 1,104 *(88.5%)* |

## 2.2 Experiment 1a: The Contribution of Climate Outliers to Predictions

This experiment determines whether historical extreme-climate cases in the PBI-CBR dataset tend to be used to predict growth rates when extreme-climate problem-cases are encountered (using 2018 as a test year).

**Setup & Method.** This experiment ran a version of PBI-CBR (for the years 2013-2016) with and without its climate outliers (as defined above); so, we compared the (a) *original* system with all training outliers *included* (PBI-CBR$^{O}$; N=36,529) and (b) PBI-CBR$^{EX}$, a version of the system with all training climate-outliers *excluded* (N=32,719). For all tests *k=30*, the value found to deliver the highest accuracy in previous tests of

[2] A *unique outlier* is a case with an extreme value on any of its weather features.

PBI-CBR [2]. The measure used was the Absolute Error (AE) found for each test case in a given year (measured in kg/DM/ha), where the *AE* = |*actual-grass-growth - predicted-grass-growth*|. *Mean Absolute Error* (MAE) is the aggregate measure over all test-cases in a given condition.

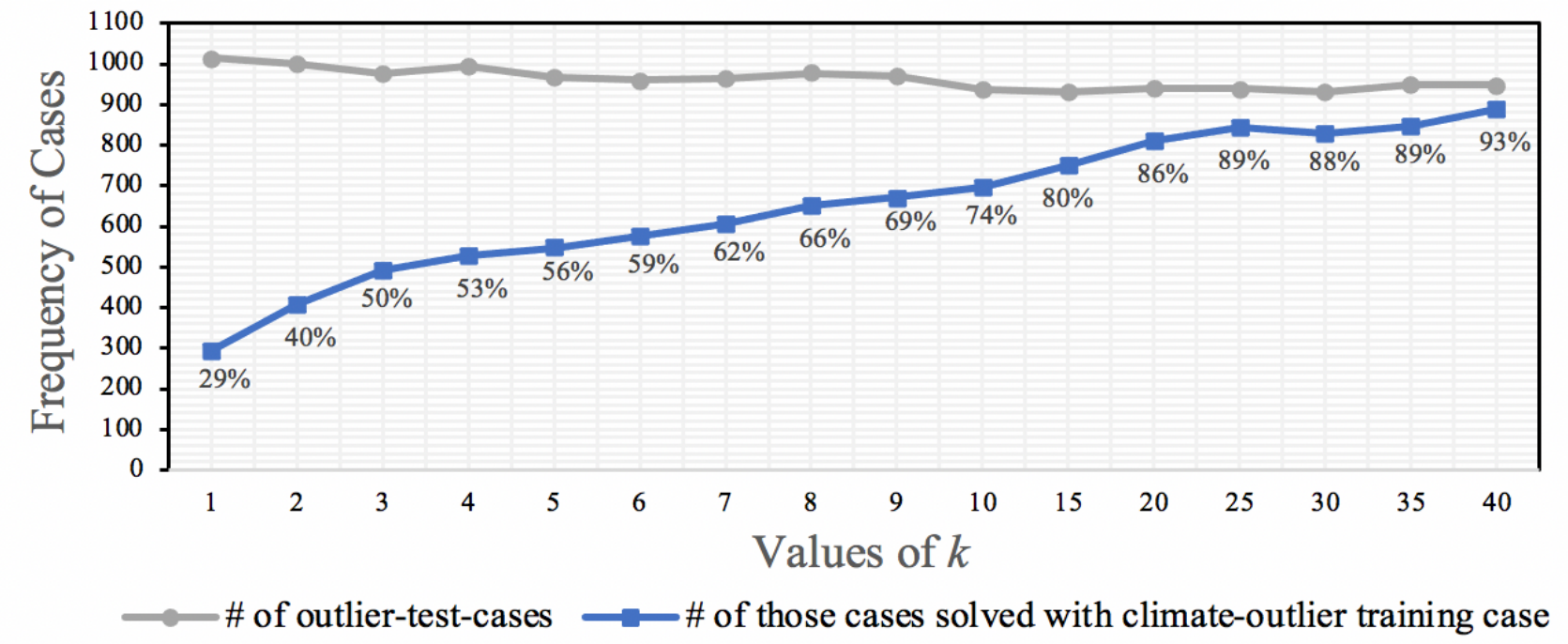

**Fig. 3.** Expt. 1b results showing the number of outlier-test-cases and frequency (and %) of outlier-test-cases solved by outlier-training-cases for different values of *k* (for 2018)

**Results & Discussion.** The results showed that the presence of climate-outliers in PBI-CBR training set significantly improved the performance of the system. The absolute-errors across the 2018 test-set showed that PBI-CBR$^{O}$ (MAE=20.55 kg/DM/ha) performed reliably better than PBI-CBR$^{EX}$ (MAE=20.63/DM/ha) which excluded the climate outliers, $t(14950) = 3.58$, $p < 0.001$, one-tailed[3]. While these MAE differences may not look large, they could be quite significant for a given farm. Remember the measure here is a kg/DM/ha (kilograms of dried grass/matter per hectare), so a 0.50 kg error could be a lot of grass, as it is multiplied by the acreage of the farm for each weekday. Importantly, we also determined which training-cases that were being used to make predictions for the 2018 test-cases to determine whether PBI-CBR$^{O}$ succeeds by using past extreme-climate events to handle new extreme-climate events. Specifically, that in PBI-CBR$^{O}$ the *climate-outliers in the training set* are used to make predictions for *climate-outliers in the test set*; note, the test-cases used for 2018, were all the outlier cases (N=1,736) in 2018 and all the non-outlier cases with "good" predictions in 2018 (N=1,248); where a *good prediction* was one with an *AE* equal to or better than the *MAE* for all test cases in that year. Table 1 shows that in solving 2018 test-cases, there is a marked tendency for *climate-outlier test-cases* to be solved by *climate-outlier training-cases* (~88% of the time). This result confirms the intuition that PBI-CBR is succeeding by flexibly assembling similar cases in atypical, local regions of the problem space to make better predictions. Recall, this result is based on *k*=30 for PBI-CBR, so in the second experiment we varied *k*, as a sensitivity analysis of this result.

---

[3] Similar results were found to tests of 2017, though less marked, as that year has fewer disruptive events: PBI-CBR$^{O}$ (MAE=18.58 kg/DM/ha) was better than PBI-CBR$^{EX}$ (MAE=18.62kg /DM/ha) which excluded the climate outliers, $t(18610) = 1.9$, $p < 0.05$, one-tailed.

### 2.3 Experiment 1b: Role of Training Outliers at Values of *k*

Expt.1a was run using the optimal *k*=30, with predictions being made by averaging the grass-growth values over all cases in *k*. However, it would be good to know at what *k*-value these training-outlier-cases begin to play a role in solving test-outlier-cases. If these training outliers appear in solving test-cases at low values of *k*, then it means these cases are being readily recruited to solve test-cases (n.b., predicted values of grass growth are based on mean of the cases in *k*). So, in this experiment *k* was varied and role of training-climate-outliers in predictions was noted.

**Setup & Method.** Using the 2018 test-set, ~1,000 test-cases with "good predictions" were tested for every value of *k* = 1-40; where a *good prediction* was one with an *AE* equal to or better than the *MAE* for all test cases in that year (n.b., differs for each *k*).

**Results & Discussion.** Figure 3 shows the results of varying *k* on the occurrence of outlier-training-cases that solve outlier-test-cases. Stated simply, it shows that by *k*=4, climate-outlier training-cases are contributing to predictions in >50% of climate-outlier test-cases showing that these key past cases are being used. So, having established that climate-outlier cases *are* used to make better predictions for disruptive climate events, in the next study we consider whether counterfactual methods can be used to generate new outlier-cases, to augment the dataset, and improve prediction even further.

## 3 Study 2: Predicting Climate Disruption with Counterfactuals

In Study 1, we saw that PBI-CBR's grass-growth predictions benefit from the use of historical extreme-climate cases to deal with future extreme-climate test-cases (answering RQ1). Notably, in unreported details, we found also that outlier cases used by PBI-CBR, to predict climate-extreme events, were sparse, *two-difference* native counterfactuals. So, these outliers typically have two feature-value differences that "change" a "normal" case into "extreme" outlier, counterfactual-cases; for instance, a normal case for farm-x in week-12 with moderate sunshine and growth is counterfactually "changed" to an outlier case with very-high solar-radiation and a very-low grass growth (as grass has been burnt off). This finding is important because it indicates the key outlier cases that are carrying the predictive load in Study 1 (and is exploited in our algorithm).

In Study 2, we determine whether counterfactual methods from XAI have a role to play in data augmentation, to answer RQ2 and RQ3. So, we explore the idea that counterfactual methods can be used to populate a case-base with new, synthetic cases that improve predictive performance. Specifically, in PBI-CBR, that counterfactual methods can find new, synthetic *outlier cases* that improve predictive performance for *extreme-climate test-cases* in the future. So, in this study, we compare the performance of PBI-CBR using its native counterfactuals as outlier cases, as a baseline, against PBI-CBR using synthetic, counterfactually-generated outlier-cases. Note, this test that pits native-counterfactual outliers against counterfactually-generated outliers to assess whether the artificial datapoints can "beat" naturally-occurring outlier cases.

Study 2 also performs comparative tests of two counterfactual algorithms from the XAI literature: Mothilal et al.'s [18] DiCE and Keane & Smyth's [4] case-based method. These two methods take quite different approaches. DiCE randomly generates a space of perturbed cases and then finds the best counterfactuals based on balancing proximity and diversity constraints [17, 18]. In contrast, the case-based method adapts cases from the original dataset; it finds a nearest neighbour to the test case, involved in a "good" counterfactual (i.e., a good native counterfactual) and then adapts the test case using the feature-differences found in this native counterfactual (see Figure 4). In the next subsection, we detail the version of this algorithm used for data augmentation; the main difference in this variation is that the counterfactual decision boundary used is not a class boundary anymore but rather the statistically-defined boundary between "normal" and "extreme" climate cases (see section 3.1, Figures 1 and 4).

### 3.1 A Case-Based Counterfactual Augmentation Algorithm (CFA)

The *Counterfactual Augmentation* (CFA) method generates synthetic counterfactual cases in three main steps: (i) "good" counterfactual pairs, $xc(x, x')$, are initially computed over the whole case-base, $X$, (ii) given a test case, $p$, a nearest neighbour case, $x$, is retrieved from the set of counterfactual pairs, $xc(x, x')$, and (iii) then, a new synthetic counterfactual case, $p'$, is produced by adapting the original test-case, $p$, using feature-difference values from $x'$. More formally:

*Definitions:*

- Normal (non-outlier) case = $x_i$ $(x1, x2, x3, \ldots, x_i)$, where $x_i \in X$
- Counterfactual (outlier) case = $x_i'$ $(x1', x2', x3', \ldots, x_i')$, where $x_i \in X$
- CF pair $xc(x, x') \Leftrightarrow target(x_i) \neq target(x_i')$
- *K*-nearest neighbors = *k*-NN
- Difference between two cases = *Diff*

**Step 1** **Identify native counterfactual (CF) pairs, $xc(x, x')$:** CFA first finds all possible "good" counterfactual pairs $xc(x, x')$ that already exist in a case-base, $X$ (pairing a normal case and its outlier counterfactual). These *native counterfactuals, xc(x, x'),* pair cases either side of *2σ climate boundary.* Each of these native pairs has a set of *match-features* and a set of *difference-features,* where the differences determine the class change (e.g., the counterfactual case may a high temperature value relative to the normal case, resulting in a different grass-growth outcome; see Figure 1).

**Step 2** **For a test case, $p$, find its nearest neighbour, $x$, from the CF pairs:** Given a test case, $p$, CFA uses *k*-NN to find its nearest neighbour, $x$, from the set of native counterfactual pairs, *xc(x, x').* The test case, $p$, is assumed to be a novel problem and, hence, not already in the case-base and therefore, does not occur in *xc(x, x').*

**Step 3** **Transfer feature values from $x'$ to $p'$ and from $p$ to $p'$.** Having identified a candidate native, *xc(x, x')* for the test case, $p$, CFA generates the synthetic counterfactual, $p'$ for $p$, such that:

- For each of the *difference-features* between $x$ and $x'$, take the values from $x'$ into the synthetic counterfactual case, $p'$.
- For each of the *match-features* between $x$ and $x'$, take the values from $p$ into the new counterfactual case, $p'$.

Clearly, the definition of a "good" counterfactual pairing is a critical parameter in this algorithm. On psychological grounds, [4] defined a "good" counterfactual to be one with no more than two feature-differences, taking a strong position on sparsity. Interestingly, subsequent user testing has shown that people prefer counterfactual explanations with 2-3 feature-differences (even over ones with 1 feature-difference [26]). Indeed, in an analysis of the outliers used in Study 1 (not reported here), we found that 2-difference native counterfactuals produced more accurate performance relative to 3-, 4- and 5-difference ones in PBI-CBR. So, the above algorithm, as in [4], uses the 2-difference definition of counterfactual "goodness" in Study 2.

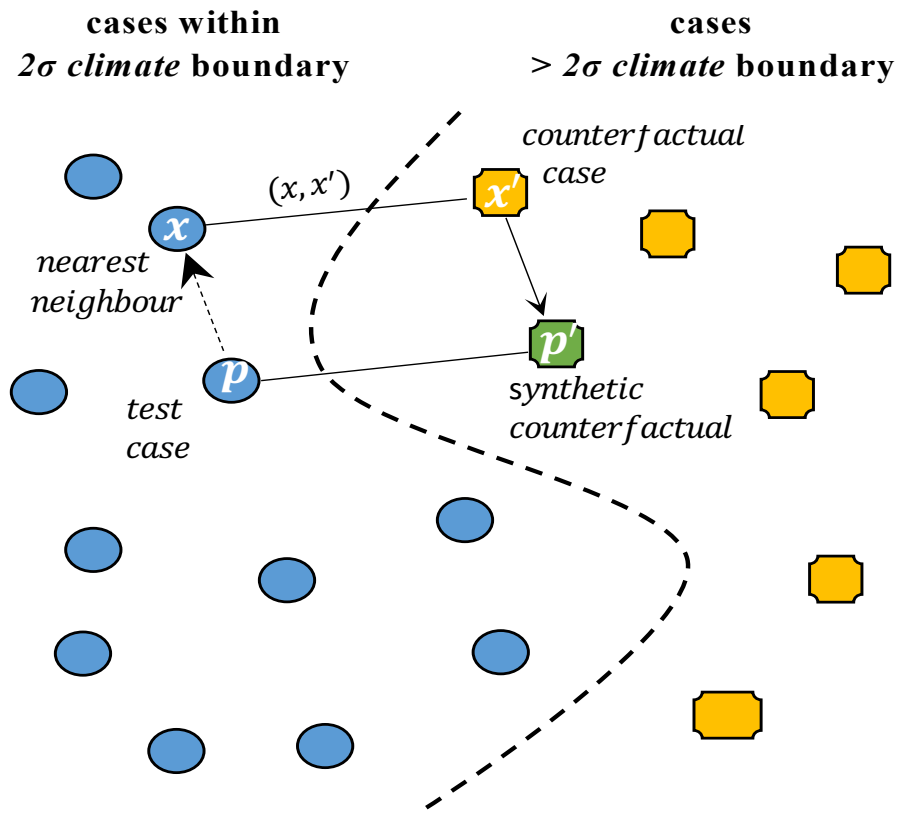

**Fig. 4**. Counterfactual Augmentation (CFA): A test case, **p**, finds a nearest neighbour, **x**, taking part in a "good" native counterfactual in the case-base, $xc(\mathbf{x}, \mathbf{x}')$, and then uses the difference-features of the counterfactual-case, $\mathbf{x}'$, to generate a new synthetic counterfactual-case, $\mathbf{p}'$, combining them with the matching-features of the original test case, **p**. The synthetic counterfactual-case, $\mathbf{p}'$, is added to the case-base to improve future prediction.

### 3.2 Experiment 2: Using Synthetic Counterfactual Cases to Predict Growth

In the present experiment, PBI-CBR's predictive performance on 2018 is run by comparing it's native-counterfactual dataset (as a baseline), against datasets of synthetic counterfactuals generated by the *Counterfactual Augmentation* (CFA) and *Diverse Counterfactual Explanations* (DiCE) methods. The interest here is specifically on how these data augmentation techniques deal with climate-extreme events. So, in these results, rather than reporting the MAE for the whole year, we break it down by month (n.b., the "year" consists of the 9 months during which cattle are grazed). In 2018, the main climate disruptions occured in March, July and October, as there was an unusually cold spring, a very hot summer, followed by a cold autumn (see Figure 5).

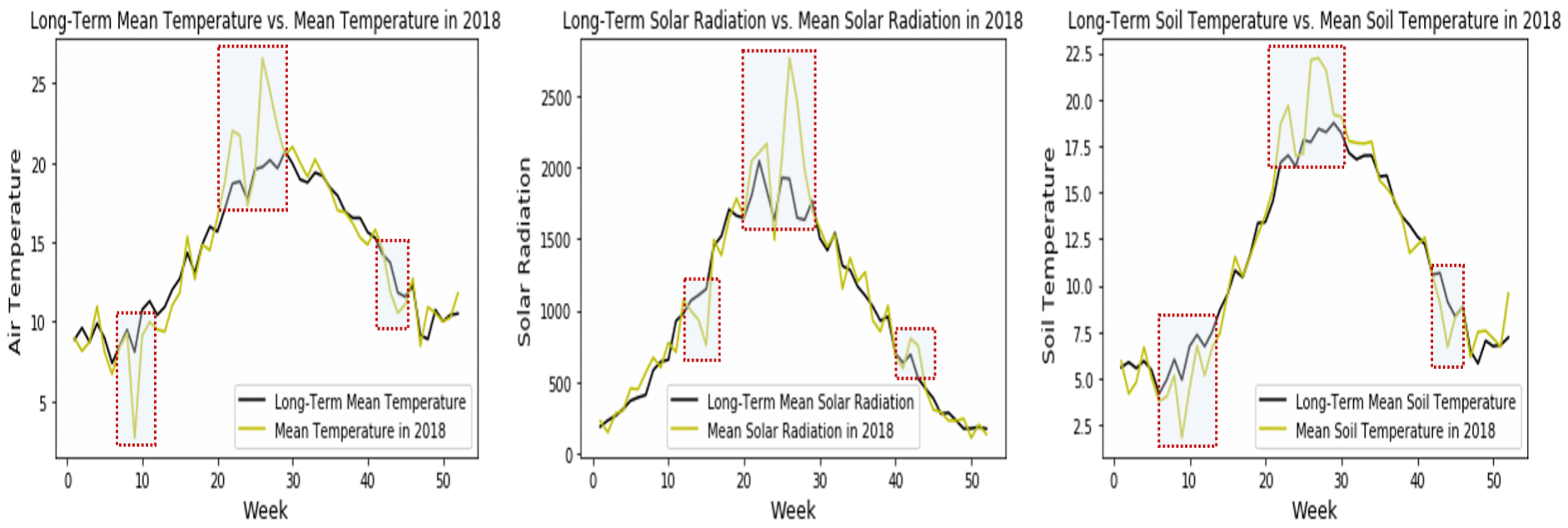

**Fig. 5.** Three graphs compare the long-term means for three weather-variables -- air temperature, solar radiation and soil temperature -- to the mean values in 2018, outlining the three main climate-disruptive periods in 2018 (i.e., March, July, and October).

**Setup & Method.** This experiment ran a version of PBI-CBR with three different datasets testing its performance against the climate-disruptive year of 2018 (see Figure 5). All datasets, used the *k*-NN to predict the grass growth-rates (measured in kg/DM/ha), using *k*=30, the value found to deliver the highest accuracy in previous tests. Again, as before, the measure used was the Mean Absolute Error (AE) found over the 2018 test-set based on averaging the *Absolute Error (AE),* where *AE* = |*actual-grass-growth - predicted-grass-growth*|. The three datasets used:

- *Native-CF:* "good" native counterfactuals (i.e., 2 feature-difference ones) from the original PBI-CBR dataset (N=2,500)
- *DiCE:* the *synthetic counterfactuals* generated by DiCE from finding the best counterfactual for the test-cases in 2013-2016 (N=2,500)
- *CFA: synthetic counterfactuals* generated by CFA based on adapting native counterfactuals for each of the test-cases in 2013-2016 (N=2,500)

Originally, we ran this experiment with unequal datasets, CFA (N=4,028) and DiCE (N=14,951) generate different numbers of counterfactuals for the 2013-2016. The current experiment equalized the counterfactual-datasets (to N=2,500) taking the mean results of 5 random case-selections. The pattern of results does not change for these test variants, but the equalized-Ns setup seems more principled.

**Table 2.** Study 2: PBI-CBR predictions (2018) for three different datasets: (i) good native counterfactuals from the original dataset (*Native-CF*), and synthetic counterfactuals from the (ii) constraint method (DiCE) and (iii) case-based method (CFA); the best results are shown in bold.

| | Mean Absolute Error (*MAE*) of growth kg DM/ha/day | | | | | | | | |
|---|---|---|---|---|---|---|---|---|---|
| | Feb | Mar | Apr | May | Jun | July | Aug | Sept | Oct |
| Native-CF | **40.9** | 40.2 | 19.0 | **26.4** | 24.8 | 30.0 | 21.7 | **16.7** | 33.0 |
| DiCE | 41.0 | 35.9 | 30.4 | 48.8 | 30.8 | 25.6 | 31.2 | 25.0 | **22.7** |
| CFA | 41.3 | **31.3** | **17.6** | 30.2 | **21.8** | **23.4** | **19.4** | 17.2 | 25.7 |

**Results & Discussion.** Table 2 shows the MAE values for the 2018 test-set for the three counterfactual datasets: Native-CF, DiCE, and CFA. Overall, the case-based method (CFA) does best in 5/9 months, with the native counterfactuals (Native-CF) doing best in 3 and DiCE just 1 (see Table 2, Figures 5 and 6); notably, CFA succeeds in periods where the most climate-disruption occurred, the cold spring (March-April), the hot summer (Jun-Aug), and is a close second to DiCE for the cold autumn (October; see Figure 6). An ANOVA tedt showed that the differences for the datasets were reliable for all months ($p < .001$). Furthermore, CFA appears to generate better data augmentations than DiCE, especially in the climate-disrupted months. The MAE across the July test-set showed that CFA (MAE=23.4 kg/DM/ha) performed better than DiCE (MAE=25.6 kg/DM/ha) and the Native-CF (MAE=30.0 kg/DM/ha) conditions, $F(2, 4446,) = 50.49$, $p < .001$; with a decrease in the error rate of up to 22%. Also, in March, the MAE decreased from 40.2 (Native-CF) to 31.3 (CFA), $F(2, 2169,) = 65.59$, $p < .001$; the more extreme the disruption the better CFA seems to perform (see Figure 6).

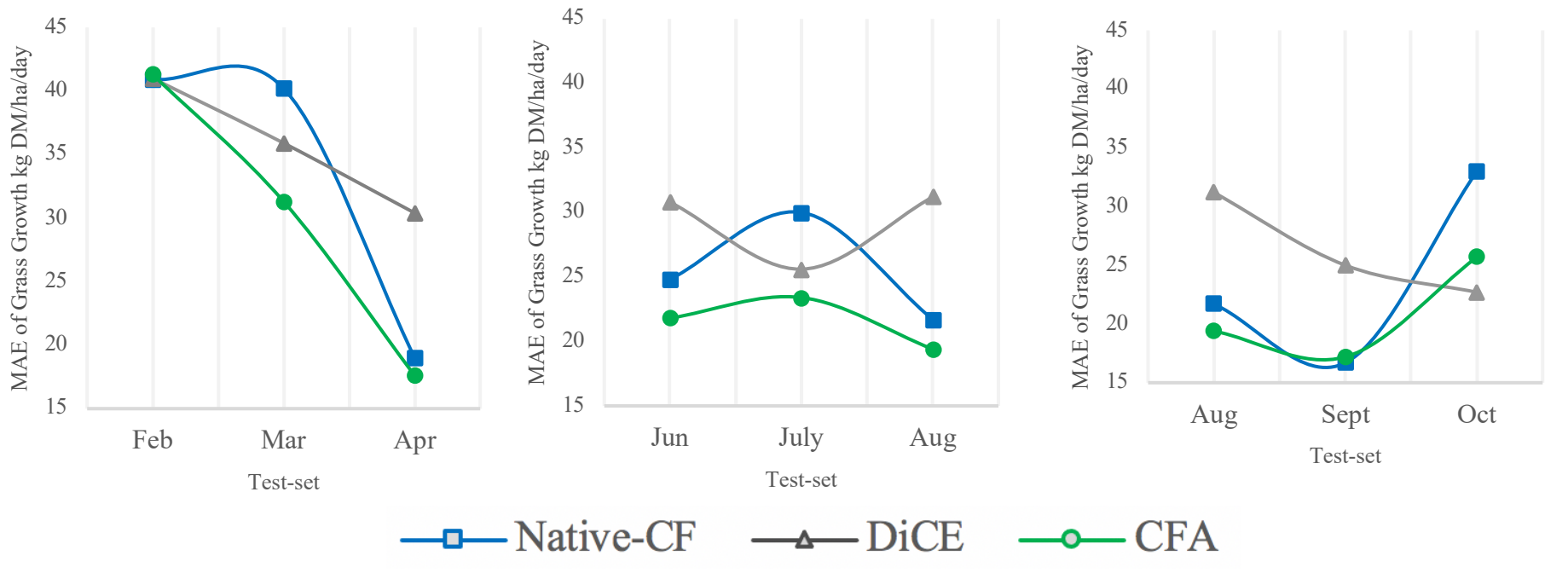

**Fig. 6.** Error in grass-growth predictions (MAE of kg DM/ha/day) in the spring, summer and autumn of 2018 for the Native-CF, CFA and DiCE datasets. Note, the counterfactual methods, CFA and DiCE, consistently do better than the native counterfactuals (Native-CF) in the climate-disrupted months (March, July, October)

## 4 Conclusions: Novelties, Explications and Caveats

The present paper exhibits some of the *promise* that AI, and specifically CBR, offers to the *challenge* of climate change; specifically, we can see how AI might be applied to climate problems in sustainable, dairy agriculture. It shows that the counterfactual methods developed for XAI can be usefully deployed to augment datasets, with synthetic cases, that improve subsequent predictions in climate-disruptive periods. This result is significant because it shows that these techniques can used to supplement historical datasets to better predict what could have been "an unpredictable future".

**Novelties.** Specifically, we have answered the three research questions posed in the introduction: we have shown that (a) the original PBI-CBR system makes accurate predictions for climate-disrupted periods by relying on historical outlier cases (RQ1), (b) its prediction of crop growth in climate-disruptive events can be improved by counterfactual data-augmentation methods (RQ2), (c) the case-based CFA method performs better on this task than a benchmark optimization method (RQ3). As such, this paper reports several significant novelties: namely, key discoveries on how (a) AI methods for data augmentation can be used to deal with climate change, (b) counterfactual methods can be successfully used for data augmentation, (c) case-based counterfactual techniques can generate useful synthetic datapoints. We conclude by considering why this counterfactual approach seems to work and what caveats/concerns need to be noted.

**Why Does This Work ?** When we first discovered these effects of counterfactual data augmentation, they appeared (to us) to be both exciting and, somewhat magical. We asked ourselves "Why does this work?". How can a generated synthetic datapoint better predict a future event over historical data? There seemed to be no good reason for why it might work? Now, having completed these experiments (and a 100 more not reported here), it is beginning to become clear why this case-based counterfactual method succeeds. CBR is often claimed to be optimal when "local" views of the data are needed to solve problems (as it seems to be here), rather than generalized, "global" functions over the whole dataset (e.g., as in iterative optimization methods, such as neural networks). When we encounter a good native-counterfactual in a dataset, we essential find a rule (a bit like an adaption rule) that tells us what minimal set of feature-changes move a case over a decision boundary. CFA exploits this implicit-knowledge in the case-base when it adapts the native-counterfactual to produce a synthetic counterfactual case, so these artificial cases are "meaningful offsets" from historical cases (it's like applying a good adaptation rule to generate new synthetic data-point). Notably, DiCE does not do this. DiCE perturbs feature-values and filters results based on broad constraints of proximity and diversity; as such, while it may "hit on" a case that is useful for solving the problem it does not do this in the guided way that CFA works. From another perspective, the present outlier cases here are essentially pivotal cases, in competence terms [21], and CFA is effectively generating novel, synthetic pivotal cases that, of course, have a high probability of being useful. These are some of the reasons why we think this case-based data augmentation approach works.

**Table 3.** PBI-CBR's growth predictions for 9 months of 2018, with (PBI-CBR$^{O+CFA}$) and without (PBI-CBR$^{O}$) the synthetic counterfactual outliers generated by the CFA method

| | Mean Absolute Error (*MAE*) of growth kg DM/ha/day | | | | | | | | |
|---|---|---|---|---|---|---|---|---|---|
| | Feb | Mar | Apr | May | Jun | July | Aug | Sept | Oct |
| PBI-CBR$^{O}$ | **22.8** | **16.7** | 17.08 | **21.2** | **23.6** | 30.3 | 19.8 | **16.5** | **16.03** |
| PBI-CBR$^{O+CFA}$ | 23.3 | 17.8 | **17.06** | 21.7 | 23.7 | **29.9** | **19.6** | 16.6 | 16.33 |

**Caveats & Concerns.** However, there are some caveats we should keep in mind about these data augmentation successes. First, we have shown these results in one dataset; so can we be confident they generalize? Temraz and Keane [27] have applied this method to many standard datasets and found similar improvements. Second, note that in Study 2 we performed a carefully controlled study, pitting native counterfactuals against synthetic ones to determine the impacts of the latter. If one was using CFA in the PBI-CBR system, one would presumably add the generated counterfactuals to the original historical dataset and then run that full-dataset on 2018. When we do this, we can see that CFA still delivers improvements, but only in the more extreme months (April, July, August; see Table 3). So, obviously, the relative impacts of these techniques will wax and wane depending on the severity of the climate events encountered. Finally, the CFA method used here could be improved: Smyth and Keane [20] have proposed a more general counterfactual method than CFA, that appears to deliver better explanatory counterfactuals. It remains to be seen whether these are also better augmenting counterfactuals. Indeed, this raises a broader question of whether the explanatory versus data-augmentation requirements on counterfactual methods will, at some stage, diverge as they do appear to be very different use-contexts. But that is, as they say, a question for another day.

## References

1. Rosenzweig, C., Iglesias, A., Yang, X.B., Epstein, P.R., and Chivian, E., Climate Change and U.S. Agriculture. centre for health and the global environment, Harvard Medical School: Boston, MA, USA, (2000)
2. Kenny, E.M., Ruelle, E., Geoghegan, A., Shalloo, L., O'Leary, M., O'Donovan, M., and Keane, M.T.: Predicting grass growth for sustainable dairy farming. In *ICCBR-19*, pp. 172-187, Springer, Berlin (2019)

3. Kenny, E.M., Ruelle, E., Geoghegan, A., Shalloo, L., O'Leary, M., O'Donovan, M., Temraz, M., and Keane, M.T.: Bayesian Case-Exclusion for Sustainable Farming. In IJCAI-20 (2020)
4. Keane, M.T. and Smyth, B.: Good counterfactuals and how to find them. In *ICCBR-20*, pages 163–178. Springer (2020)
5. EU Parliament : Briefing on the EU dairy sector. https://www.europarl.europa.eu/RegData/etudes/BRIE/2018/630345/EPRS_BRI(2018)630345_EN.pdf (2018)
6. Altieri, M.A.: *Agroecology: the science of sustainable agriculture*. CRC Press (2018).
7. Teagasc: The Dairy Carbon Navigator: Improving carbon efficiency on Irish dairy farms.
8. Ruelle, E., Hennessy, D. and Delaby, L.: Development of the Moorepark St Gilles grass growth model (MoSt GG model). *European Journal of Agronomy*, **99**, pp.80-91 (2018)
9. Hanrahan, L., Geoghegan, A., O'Donovan, M., Griffith, V., Ruelle, E., Wallace, M. and Shalloo, L.: PastureBase Ireland. *Computers and Electronics in Agriculture*, *136*, 193-201 (2017)
10. Hurtado-Uria, C., Hennessy, D., Shalloo, L., O'Connor, D. and Delaby, L.: Relationships between meteorological data and grass growth over time in the south of Ireland. Irish Geography, **46**(3), 175-201 (2013)
11. Karimi, A.H., Barthe, G., Schölkopf, B. and Valera, I.: A survey of algorithmic recourse: definitions, formulations, solutions, and prospects. arXiv preprint arXiv:2010.04050 (2020)
12. Keane, M.T., Kenny, E.M., Delaney, E. and Smyth, B.: If Only We Had Better Counterfactual Explanations. arXiv preprint arXiv:2103.01035 (2021)
13. Dodge, J., Liao, Q.V., Zhang, Y., Bellamy, R.K. and Dugan, C.: Explaining models. In IUI-19, pages 275-285 (2019)
14. Nugent, C., Doyle, D. and Cunningham, P.: Gaining insight through case-based explanation. Journal of Intelligent Information Systems, **32**(3), pages 267-295 (2009)
15. McKenna, E. and Smyth, B.: Competence-guided case-base editing techniques. In European Workshop on Advances in Case-Based Reasoning, pages 186-197, Springer, Berlin (2000)
16. Dasarathy, B.V.: Minimal consistent set (MCS) identification for optimal nearest neighbor decision systems design. IEEE Transactions on Systems, Man, and Cybernetics, **24**(3), pages 511-517 (1994)
17. Wachter, S., Mittelstadt, B. and Russell, C.: Counterfactual explanations without opening the black box: Automated decisions and the GDPR. Harv. JL & Tech., **31**, p.841 (2018)
18. Mothilal, R.K., Sharma, A. and Tan, C.: Explaining machine learning classifiers through diverse counterfactual explanations. In FAT*20, pages 607-617 (2020)
19. Schleich, M., Geng, Z., Zhang, Y. and Suciu, D.: GeCo: Quality counterfactual explanations in real time. arXiv preprint arXiv:2101.01292 (2021)
20. Smyth, B. and Keane, M.T.: A few good counterfactuals. *arXiv prepring:2101.09056* (2021)
21. Smyth, B. and Keane, M.T.: Remembering to forget. In: Proceedings of the 14th international Joint Conference on Artificial intelligence (IJCAI-95) pp. 377-382 (1995).
22. Hasan, M.G.M.M., Use case of counterfactual examples: Data augmentation. Proceedings of Student Research and Creative Inquiry Day (2020)
23. Subbaswamy, A. and Saria, S.: Counterfactual normalization: Proactively addressing dataset shift using causal mechanisms. In UAI-18, pp. 947-957 (2018)
24. Zeng, X., Li, Y., Zhai, Y., and Zhang, Y.: Counterfactual Generator In: Proceedings of the 2020 Conference on Empirical Methods in Natural Language Processing, 7270–7280 (2020)
25. Pitis, S., Creager, E., and Garg, A.: Counterfactual data augmentation using locally factored dynamics. Advances in Neural Information Processing Systems (2020)
26. Förster, M., Klier, M., Kluge, K. and Sigler, I.: Fostering Human Agency: A Process for the Design of User-Centric XAI Systems. In ICIS-2020, paper 1963 (2020).
27. Temraz, M. & Keane, M.T.: Using counterfacutals for handling the class imbalance problem, unpublished report, March 2021.